\pdfoutput=1

\PassOptionsToPackage{dvipsnames,usenames}{xcolor}
\PassOptionsToPackage{hyphens}{url}

\documentclass[11pt,table,xcdraw]{article}

\usepackage{acl}

\usepackage[usenames,dvipsnames]{xcolor} 
\usepackage{latexsym}
\usepackage{graphicx}
\usepackage{tabularx}

\usepackage{booktabs}
\usepackage{amsmath}
\usepackage{amsfonts}
\usepackage[varg]{txfonts} 
\usepackage{setspace} 
\usepackage[hyphens]{url}
\usepackage{xspace}
\usepackage{multirow}
\usepackage{makecell}

\usepackage{hyperref}
\usepackage{graphicx}
\usepackage{fvextra}

\usepackage{listings}

\usepackage{graphicx}
\usepackage{caption}  
\usepackage{svg}
\usepackage{enumitem}

\usepackage[T1]{fontenc}

\usepackage[utf8]{inputenc}
\usepackage{csquotes}

\usepackage{microtype}
\usepackage{silence}
\usepackage{inconsolata}

\title{EmPO: Emotion Grounding for Empathetic Response Generation through Preference Optimization}

\author{Ondrej Sotolar \\ {\bf Vojtech Formanek}\\ Masaryk University, Brno \\ xsotolar@fi.muni.cz \And
        Alok Debnath \\ Trinity College, Dublin \And
        Allison Lahnala \\ {\bf Charles Welch} \\ {\bf Lucie Flek} \\ CAISA, University of Bonn}

\begin{document}
\maketitle
\begin{abstract}
Empathetic response generation is a desirable aspect of conversational agents, crucial for facilitating engaging and emotionally intelligent multi-turn conversations between humans and machines. Leveraging large language models for this task has shown promising results, yet challenges persist in ensuring both the empathetic quality of the responses and retention of the generalization performance of the models. We propose a novel approach where we construct theory-driven preference datasets based on emotion grounding and use them to align LLMs with preference optimization algorithms to address these challenges. To evaluate empathetic response generation, we employ the \textsc{EmpatheticDialogues} dataset, assessing empathy with the \textsc{diff-Epitome} and BERTscore metrics and with multi-dimensional human evaluation. Additionally, we measure diversity and emotional valence using feature-based methods. We also evaluate the impact of training on the generalization performance using the MMLU benchmark and tasks from the Open LLM Leaderboard. The results show that LLMs can be aligned for empathetic response generation by preference optimization while retaining their general performance and that emotion grounding can guide preference dataset creation. 
\end{abstract}

\section{Introduction}
\label{sec:intro}
Empathetic response generation (ERG) focuses on tuning a conversational agent toward understanding the user's situation, feelings, and experience to generate appropriate, human-like responses. This goal was first attempted by the earliest rule-based chatbots like ELIZA \citep{weizenbaum1966eliza}, then more recently with deep learning approaches such as used by the chatbot BlenderBot \citep{roller2021recipes,xu2022beyond,komeili2022internet,shuster2022blenderbot}. The introduction of instruction-tuned large language models (LLMs) such as ChatGPT \cite{OpenAI2023ChatGPT} has challenged the contemporary approaches to ERG. These models can participate in multi-turn conversations without additional training and are implicitly capable of generating empathetic conversations \citep{lee-etal-2022-gpt}. 

LLMs also have shown tremendous generalization capabilities and can draw on world knowledge obtained by training on Internet-size data. These traits were not present in the earlier non-LLM conversational agents. Recent developments focus on approaches that retain the generalization abilities of LLMs while improving empathy in responses, including prompt-engineering methods that do not change the model's weights but rather tune the prompt content to assist the generation by in-context learning~\citep{wang2024sibyl}, often assisted by retrieval~\citep{qian-etal-2023-harnessing}. 

In the current study, we propose a data-driven solution for ERG with LLMs: aligning LLMs via preference optimization algorithms. First, we build a preference dataset using the benchmark dataset \textsc{EmpatheticDialogues}~\citep{rashkin-etal-2019-towards}. It contains short multi-turn human-to-human dialogues grounded by emotion labels. We leverage this emotion grounding to sample dialog completions labeled with polar opposite emotions using Plutchik's wheel \citep{plutchik2001nature} such that each prompt is paired with preferred and non-preferred completions. We then fine-tune a foundational LLM using Direct Preference Optimization (DPO)~\citep{rafailov2024direct} to generate responses aligned with the preferred candidate response.

Using metrics such as the model-based empathy metric for multi-turn conversations \textsc{diff-Epitome}~\citep{lee2022does}, several feature-based metrics for diversity and emotional valence, and multi-dimensional human evaluation, we show that training LLMs with our preference dataset helps to generate diverse and empathetic responses. We also investigate the models' general language understanding using benchmarks such as the Massive Multitask Language Understanding (MMLU)~\citep{hendrycks2020measuring} and the tasks from the Open LLM Leaderboard~\citep{open-llm-leaderboard}. By monitoring how the training impacts language understanding performance, we show that it is possible to find hyperparameters that align the model for ERG while preventing catastrophic forgetting. Our method is analogous to providing guardrails with helpful/harmful preference datasets~\citep{bai2022training}. Models aligned in this way can be further adapted for any task by contemporary prompt-engineering methods or additional training which is an advantage over alignment with methods that do not train the model weights. We also share novel observations from searching over the hyperparameter configuration space and provide code to apply our method to other datasets and models.
We make all datasets, source code, and models publicly available.\footnote{\url{github.com/justtherightsize/empo}}


\section{Related Work}
\label{sec:relatedwork}
Systems specialized for ERG such as KEMP~\citep{li-etal-2022-kemp}, CEM~\citep{CEM2021}, MIME~\citep{majumder-etal-2020-mime}, EmpDG~\citep{li-etal-2020-empdg}, and \textsc{DiffusEmp}~\citep{bi-etal-2023-diffusemp} or universal conversational agents like BlenderBot were not built for general language understanding or instruction-following. They can be evaluated on dialog completion, but they fail on other tasks as opposed to LLMs; thus, they are not solutions to the same problem. 

Recent approaches to ERG such as \citet{wang2024sibyl,qian-etal-2023-harnessing,li2024enhancing} employ LLMs enhanced by prompt engineering with some success. However, their approach differs principally from ours; we restructure the problem of ERG as an LLM alignment problem and produce a method for creating preference datasets and a modeling approach rather than a prompt construction algorithm. 

To our knowledge, the current study is the first to explore aligning LLMs for ERG via preference optimization.

\paragraph{Datasets}
\label{sec:datasets}
Several dialogue datasets, including IEMOCAP~\citep{busso2008iemocap}, MELD~\citep{poria-etal-2019-meld}, DailyDialog~\citep{li-etal-2017-dailydialog}, \citealp{herzig2016classifying}, EmotionLines~\citep{hsu-etal-2018-emotionlines}, EmpathicReactions~\citep{buechel-etal-2018-modeling}, EmoContext~\citep{chatterjee-etal-2019-semeval}, and ESConv~\citep{liu-etal-2021-towards} contain emotion labels. However, these datasets are either labeled with only a small set of emotions, limited by size, or lacking a multi-turn character.

\textsc{EmpatheticDialogues} (ED) is a dataset with 25K human-to-human dialogues and 32 emotion labels, which are derived from biological responses~\citep{ekman1992there,plutchik1980general} to larger sets of subtle emotions derived from contextual situations~\citep{skerry2015neural}. It has become the benchmark dataset on empathetic conversation because it addresses the limitations of the preceding datasets. 

Earlier deep learning-based conversational agents required large datasets for training from scratch to learn to participate in dialogue, often at the expense of the dataset quality. An example is the EDOS dataset~\citep{welivita2021large} of 1M movie dialogues with emotion grounding annotated by classifiers trained on ED.

However, ED and its successors have been criticized for their data annotation and model evaluation approaches \citep{debnath-2023-critical}. In treating ERG as an LLM alignment task, our approach addresses some of these drawbacks.

Recently, it became feasible to generate high-quality synthetic datasets using the largest LLMs, such as the Chinese SMILE \citep{qiu2024smile} and Korean SoulChat \citep{chen2023soulchat} datasets. The respective works demonstrate that synthetic datasets can be successfully used for training LLMs to improve in ERG. However, neither study evaluated the general language understanding of the trained models, so it is impossible to ascertain whether the training led to the specialization of the models. 

\paragraph{LLM Alignment by Preference Optimization}
Aligning generative models with human feedback has improved their helpfulness, factual accuracy, and ethical behavior, among other aspects~\cite {ouyang2022training}. Methods like RLHF~\citep{christiano2017deep}, including training algorithms such as PPO~\citep{schulman2017proximal} and DPO~\citep{rafailov2023direct} have consistently been more effective than relying solely on supervised fine-tuning (SFT). Human feedback can take many forms: PPO requires human preferences to rank the generations, and DPO requires datasets of prompts and pairs of preferred/rejected completions. 

\paragraph{Automated Evaluation of Empathy in Dialogue}
Evaluation of empathy in dialogue has primarily focused on human evaluation and lexical overlap such as BLEU~\citep{papineni2002bleu} or ROUGE~\citep{lin2004rouge} or semantic similarity with reference dialogues such as BERTscore~\cite{zhang2019bertscore}. The former is widely used. However, it is limited by the subjective nature of the task, biases introduced by the evaluation design, and, usually, small sample sizes. Given the highly fluent nature of generative LLM responses, we have found the latter to have little relevance for measuring empathy, in agreement with the findings of~\citealp{liu2016not}.
More recent methods rely on model-based empathy metrics, which use classification models trained specifically for measuring empathy in multi-turn conversations such as \textsc{diff-Epitome}~\citep{lee2022does}. \textsc{diff-Epitome} is an evolution of the \textsc{Epitome} classifiers~\citep{sharma2020computational}, which measure empathy in dialogue on a scale (0-2) from none (0), through weak (1), to strong (2) on three dimensions: empathetic responses (ER), explanations (EX), and interpretations (IP). \textsc{diff-Epitome} uses a similar but continuous scale on the same dimensions by averaging the scores across the entire dataset, thus providing a measure of difference from the ground truth.

In contrast to representation-based evaluation models, \citet{lee-etal-2024-comparative} employ feature-based methods for evaluation. These include specificity based on a normalized variant of inverse document frequency (NIDF,~\citet{see-etal-2019-makes}), which shows emotional connections and language that conveys an understanding of the other's feelings and experiences \cite{truax1964concreteness}. It further includes word choice metrics derived from the NRC Emotion Intensity Lexicon \cite{mohammad-2018-word}, which are intended to capture the consistency of the response with the dialog context; thus, valence, arousal, and intensity are measured for both the prompt and response and the ideal is for them to score similarly. Finally, response diversity is measured using response-tries as consideration of the common pitfall of generation systems to resort to generic responses like ``I am so sorry to hear that'' \cite{lee-etal-2024-comparative}. The work also offers an averaging schema to account for the dialog context.

Reflectivity has also been used as a proxy measure for empathy: \citet{lee-etal-2024-comparative} used PAIR~\cite{min-etal-2022-pair} to measure the reflection level in empathetic utterances. 

\paragraph{Human Evaluation of Empathy in Dialogue}
Human judges, usually crowd workers or students, have been extensively used to evaluate empathy in dialogue. The judges are instructed to evaluate the generated responses in comparative A/B tests or one model at a time in a scale-based evaluation. For the former, an alternative yes/tie/no rating is common, and the result is derived by majority voting. The latter employs three or more dimensions, such as \textit{empathy}, \textit{relevance}, \textit{informativeness}, \textit{coherence}, or \textit{fluency}. The quality of the surveys varies as some studies apply one question per dimension, and the exact description of the physical surveys is rare. The sample size is usually 100-150 samples per model. Some stratify the sample on the grounding emotions such as \textsc{MIME}~\citep{majumder2020mime}. We present an overview of the evaluation methods used with \textsc{EmpatheticDialogues} in Table~\ref{tab:rw-eval}.
\begin{table*}[ht]
  \centering
  \small
  \begin{tabular}{l|l|l}
    \hline
    \textbf{Study} & \textbf{Type} & \textbf{Dimensions}  \\\hline
     \textsc{EmpatheticDialogues}~\citep{rashkin-etal-2019-towards}& \multirow{4}{*}{A/B, 5p Likert} & \multirow{5}{*}{empathy, relevance, fluency}\\
     MoEL~\citep{lin-etal-2019-moel}&&\\
     MIME~\citep{majumder2020mime}&&\\
     EmpDG~\citep{li-etal-2020-empdg}&&\\\cline{1-2}
     EmoCause/GEE~\citep{kim-etal-2021-perspective} &A/B, 4p Likert&\\\hline
     ESConv~\citep{liu-etal-2021-towards}& Likert &fluency, identification,\\
     &&comforting, suggestion\\\hline
     EmpBot~\citep{zaranis2021empbott5basedempatheticchatbot}&\multirow{2}{*}{5p Likert}& fluency \& relevance,\\
     &&empathy-sentiment,\\
     &&empathy-emotion\\\hline
     CARE~\citep{wang-etal-2022-causal} & \multirow{2}{*}{A/B, 5p Likert} & \multirow{3}{*}{coherence, empathy, informativeness}\\
     \textsc{DiffusEmp}~\citep{bi-etal-2023-diffusemp} & &\\\cline{1-2}
     CEM~\citep{CEM2021}&aspect-wise A/B&\\\hline
     EPITOME~\citep{lee2022does}& \multirow{2}{*}{3p Likert} &emotional reactions, interpretations,\\
     &&explorations\\\cline{1-1}\cline{3-3}
     SoulChat~\citep{chen2023soulchat}&& content, empathy, helpfulness, safety\\\hline
  \end{tabular}
  \caption{Overview of human evaluation methods in recent related work.}
  \label{tab:rw-eval}
\end{table*}

Human evaluation offers valuable gold-standard feedback but not without drawbacks, including subjectivity, limited sample size, and the challenge of designing bias-free surveys. LLM-as-a-judge methods have emerged to address these issues. However, their validity remains contested because surveys such as \citet{bavaresco2024llmsinsteadhumanjudges} point out that these methods introduce several new biases, including the position bias~\citep{koo2024benchmarkingcognitivebiaseslarge}, form-over-substance~\citep{wu2023stylesubstanceevaluationbiases}, model-relatedness~\citep{baris-schlicht-etal-2024-pitfalls}, and length sensitivity~\citep{chen2024humansllmsjudgestudy}.

\paragraph{Language Understanding Evaluation} 
An invaluable trait of LLMs is their ability to generalize across various domains, tasks, and languages. Language understanding evaluation compares and ranks different models based on this ability. From the many existing benchmarks, two types are especially relevant for dialog-capable models: live A/B testing using either human judgments such as Chat Arena~\citep{chiang2024chatbot} or using LLM-as-a-judge such as in AlpacaEval~\citep{dubois2024lengthcontrolledalpacaevalsimpleway} and static language understanding benchmarks such the MMLU or the Open LLM leaderboard tasks.

Static benchmarks offer consistency and larger sample sizes. MMLU assesses how instruction-following LLMs learn and apply knowledge across various domains by measuring their performance on 16K different multiple-choice style question tasks from STEM, humanities, and world knowledge. It shows the best correlation with human judgments from the Chat Arena among large static benchmarks~\cite{dubois2024lengthcontrolledalpacaevalsimpleway}. The Open LLM leaderboard~\cite{open-llm-leaderboard} is a popular task aggregator to evaluate and rank LLMs. It comprises six tasks that cover the ability to follow instructions, math, science, language understanding (e.g., sarcasm detection, name disambiguation), world knowledge, and multistep reasoning. See Appendix~\ref{appx:openllm} for a full description of the tasks.

\section{Experimental Setup}
\label{sec:experiments}
We conduct our experiments using the Zephyr-7B~\citep{tunstall2023zephyr} model, a variant of the Mistral-7B foundational LLM~\citep{jiang2023mistral} fine-tuned for multi-turn dialogues using the UltraChat dataset~\citep{ding2023enhancing}. \citet{open-llm-leaderboard} shows that the smallest variant with 7 billion parameters already provides good general language understanding ability as it matches LLama-2-70B on many benchmarks. Using our preference dataset, we compare the untrained baseline Zephyr model to the model trained with DPO.  We explore the hyperparameter configuration space for all training steps to minimize the impact on generalization while improving empathy. For control, we added two other sources of responses: the ground-truth responses from ED and the baseline Zephyr model, where we limited the generated length to be similar to the ground-truth. This allows us to offset responders' cognitive bias towards longer responses being considered of higher subjective quality \citep{santhanam2020studying,park2024disentangling}.

We evaluate all automated metrics using the pre-defined test set of \textsc{EmpatheticDialogues} of 2,540 dialogues.

\paragraph{Supervised Fine-tuning}
\citet{tunstall2023zephyr} showed that SFT is a necessary first step in alignment, ensuring the preference dataset is in-domain for the aligned model. We perform SFT using the standard causal auto-regressive objective using the individual dialogues from \textsc{EmpatheticDialogues}. As per the definition of the dataset, the odd turns are considered the "user prompts" and the even turns the "responses". We limit the SFT training to the even-indexed turns by masking the odd turns to ignore them in the loss-function computation. For computational efficiency, we fine-tune using LoRA adapters~\citep{hu2021lora}. 

\paragraph{Preference Optimization}
For DPO, we build a preference dataset from the \textsc{EmpatheticDialogues} consisting of preferred/rejected completions. For each dialogue, we target the last even turn (the last "response" to user) as the generation target while including the previous turns as context. This is the standard way of processing the dataset, also done in previous works. 

For constructing the completion pairs, we leverage a property of the ED dataset: each dialogue is associated with an emotion label. These were used as grounding during the ideation of each dialogue. For the preferred completion, we use the ground truth -- the original response. For its rejected counterpart, we use Plutchik's wheel of emotions (Figure~\ref{fig:plutchik}) and the derivative emotional dyads (Figure~\ref{fig:dyads}, \citealp{plutchik2001nature}) to find the polar opposite emotion labels resulting in a lookup table (see Appx.~\ref{appx:plutchik}). For each completion, we randomly select one from the group of completions labeled with the opposite label. Because the training stability can suffer from random selection, we draw a fresh random completion for each new training epoch and search for the hyperparameter configuration to offset the repetition of the preferred completions. 

\begin{table*}[ht]
  \centering
  \small
  \begin{tabular}{l|l|lll|l}
    \hline
    \textbf{Model}              &$\mathbf{\overline{x}}($\textbf{\textsc{diff-Epitome}}$)\downarrow$& \textbf{diff-ER $\downarrow$}& \textbf{diff-EX $\downarrow$}& \textbf{diff-IP $\downarrow$}& \textbf{FBert $\uparrow$}  \\\hline
    \textit{baseline:} Zephyr-7B-sft-full&  $1.055\pm.00$&$1.120\pm.00$  &$1.441\pm.00$           &$\mathbf{0.603}\pm.00$          &$0.860\pm.00$         \\
    \textit{length-controlled baseline}&    $0.786\pm.00$&$1.090\pm.00$  &$\mathbf{0.660}\pm.00$  &$\mathbf{0.613}\pm.00$          &$0.865\pm.00$         \\
    Zephyr-7B + SFT              &          $0.773\pm.00$&$0.945\pm.00$          &$\mathbf{0.647}\pm.00$           &$0.726\pm.00$          &$\mathbf{0.870}\pm.00$  \\
    Zephyr-7B + SFT + DPO        &          $\mathbf{0.701}\pm.01$&$\mathbf{0.588}\pm.01$ &$0.720\pm.01$  &$0.796\pm.01$ &$0.868\pm.00$  \\\hline
  \end{tabular}
  \caption{Model-based empathy metrics \textsc{diff-Epitome} and semantic similarity metric BERTscore with \textsc{EmpatheticDialogues}. All presented results are means ($\pm$ SD) of scores over multiple (4) training runs  with the same hyperparameters but different seeds (and different random completion draw for DPO). Bold indicates best results and significant difference ($p < $1e-5) to the next best result using permutation tests.}
  \label{tab:results}
\end{table*}

\begin{table}[ht]
  \centering
  \small
  \begin{tabular}{l|p{1.5cm}p{1.5cm}}
    \hline
    \textbf{Model}                       & \textbf{MMLU (5s) $\uparrow$}&\textbf{Open LLM (Acc) $\uparrow$} \\\hline
    \textit{baseline:} Zephyr-7B-sft-full&$.588\pm.00$                  &$.276\pm.00$\\
    Zephyr-7B + SFT                      &$.585\pm.00$                  &$.270\pm.00$\\
    Zephyr-7B + SFT + DPO                &$.584\pm.00$                  &$.266\pm.00$\\\hline
  \end{tabular}
  \caption{Language understanding metrics: MMLU and Open LLM Leaderboard accuracy over all its tasks. All presented results are means ($\pm$ SD) of scores over multiple (4) training runs  with the same hyperparameters but different seeds (and different random completion draw for DPO). With the SFT (\texttt{lr=1-e5, $\alpha$=64, epoch=1,  batch\_size=64, rank=1024}) and DPO (\texttt{lr=1e-6, $\beta$=0.05, epoch=3}) hyperparameters, we didn't find a significant difference using the McNemar's test and SD in all cases was ($ < 0.001$).}
  \label{tab:results2}
\end{table}

\paragraph{Automated Empathy Evaluation} For measuring empathy, we use the \textsc{diff-Epitome} metric, which measures the difference between levels of empathy of the ground truth and generations. We consider a lower difference from the ground truth as better. We used \textsc{diff-Epitome} also for model selection during the fine-tuning using a validation sample of the training data. 

In addition, we consider the set of feature-based metrics from \citet{lee-etal-2024-comparative} for further analysis on the generations for the test set, which include NIDF for specificity, lexicon-based valence-arousal-intensity scores (VAD), and diversity metric based on response-tries. The VAD metrics measure the difference between the prompt's and response's scores, smaller differences being the desired outcome. 
To reflect this, we consider the score distance, i.e., the absolute value of the difference, before aggregating the results per prompt-response pair (whereas the prior work aggregated the prompt score minus the response score).

We excluded common lexical-overlap and vector-similarity metrics because our preliminary experiments found them uncorrelated with the perceived quality of the generated responses. It was especially evident with the largest LLMs, such as GPT4~\citep{achiam2023gpt} and Claude~\citep{introducing-claude}, which generated high-quality responses yet received low scores from overlap metrics. However, we keep the semantic similarity BERTscore as a sanity check and for increased metric diversity. Furthermore, we could not reproduce the PAIR model, which is included among \citet{lee-etal-2024-comparative}'s metrics.


\paragraph{Language Understanding Evaluation} We measure the general language understanding of the models with MMLU and tasks from the Open LLM leaderboard using the implementation in \texttt{lm-evaluation-harness}~\cite{eval-harness}.


\paragraph{Human Evaluation}
Taking into account critique of human evaluation \citep{lahnala-etal-2022-critical, lee-etal-2024-comparative}, and building off of the lack of specificity described in \citet{Coll2017-lv}, we adapted the existing \textsc{EmpatheticDialogues} singular question about situational empathy into four dimensions: \textit{emotion understanding}, \textit{situational appropriateness}, \textit{response naturalness}, and \textit{conversational engagingness}.
Non-empathy associated metrics were combined into a single question of \textit{understandability and consistency}. The evaluators were hired using the Prolific platform \cite{prolific}
and screened for education in Psychology or Social Work.

The evaluations of the four empathy dimensions were performed on a continuous scale from 0-100 where evaluators were asked to rate their degree of agreement to Likert-like statements for each dimension \citep{ji-etal-2022-achieving}. Understandability and consistency was evaluated with a checkbox. The evaluation sample comprises 135 examples, each of the 32 emotions in ED represented at least four times. Each example was evaluated by three annotators in batches of 15 examples which were randomized to contain blocks of outputs from different models. In total 108 batches (1620 samples) were evaluated. Sample averaged annotator ratings for each dimension and model, which are averaged, form the final score. We used the pairwise t-test and Cohen's \textit{d} on $\alpha=0.05$ to test the difference between different response sources. See Appendix~\ref{appx:HE} for more details on the experiment design.

\section{Results}
\label{sec: results}

\subsection{Automatic Evaluation}
Tables~\ref{tab:results} and~\ref{tab:results2} compare the baseline, unaligned model to the supervised fine-tuned model (SFT) and to the SFT model further aligned with a preference optimization algorithm (DPO) on the main empathy and language understanding metrics. 

\begin{figure}
    \centering
    \includegraphics[width=\linewidth]{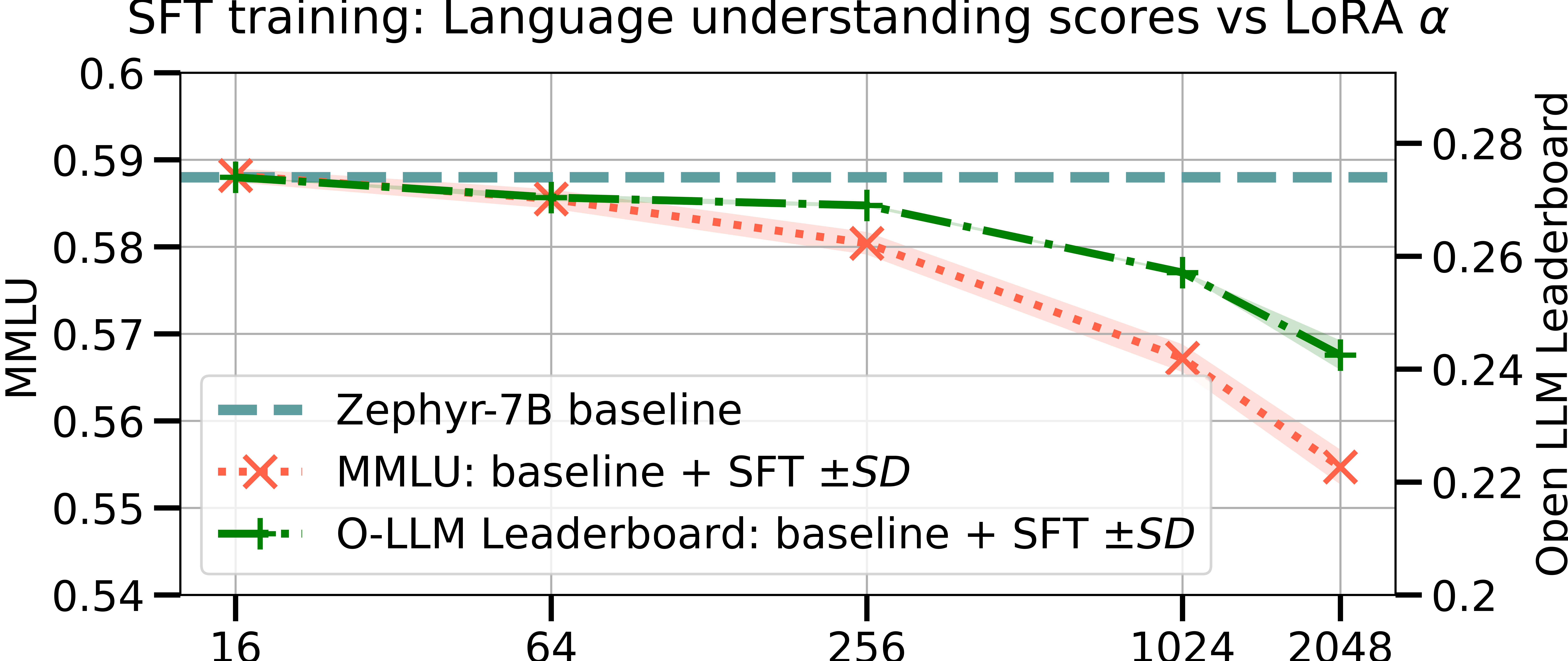}
    \caption{Supervised fine-tuning with LoRA: impact of the hyperparameter $\alpha$ on the MMLU score and Open LLM Leaderboard accuracy. Trained with: \texttt{lr=1-e5, batch\_size=64, rank=1024}.}
    \label{fig:mmluvsalpha}
\end{figure}

\begin{figure}
    \centering
    \includegraphics[width=\linewidth]{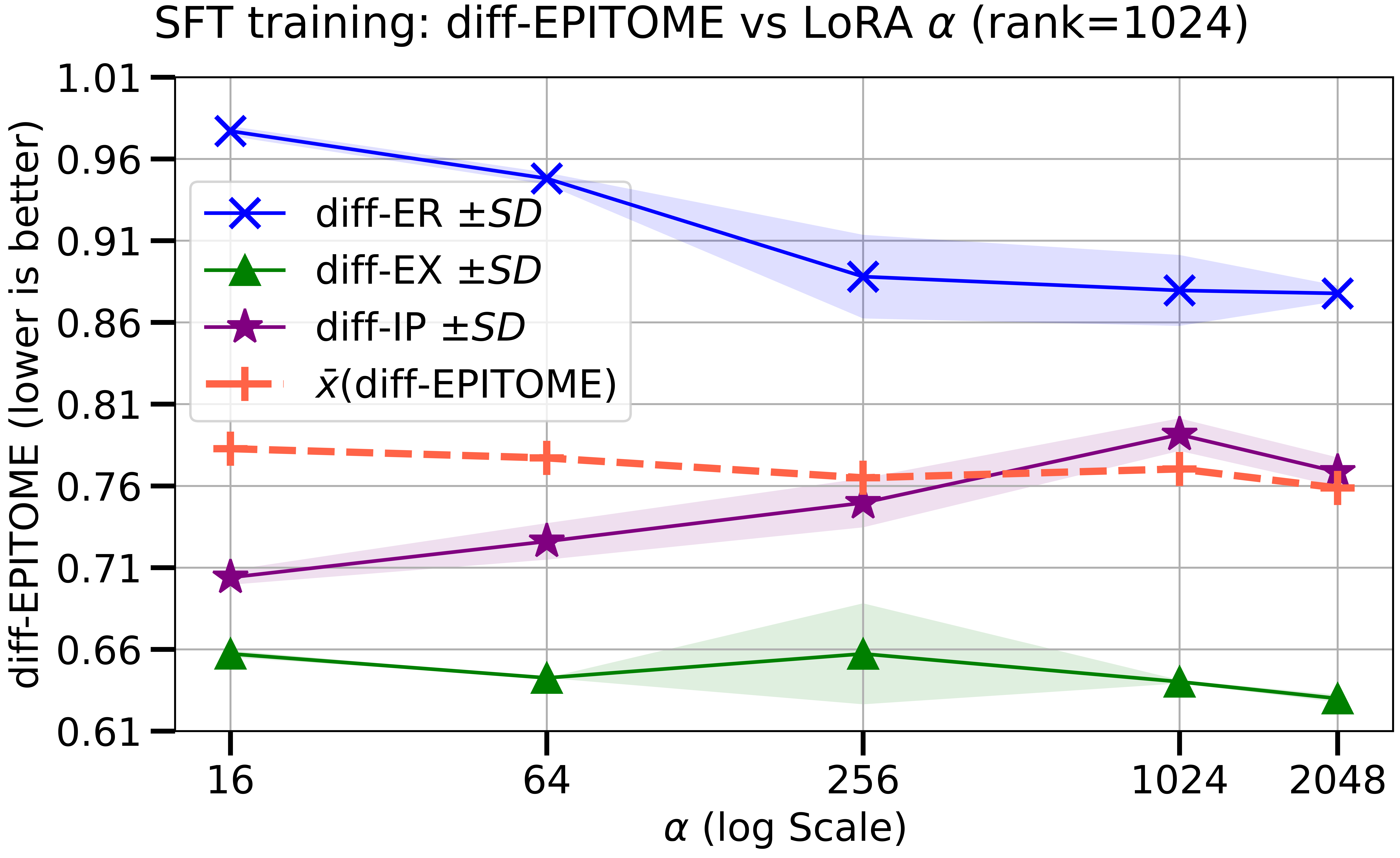}
    \caption{Supervised fine-tuning with LoRA: impact of the hyperparameter $\alpha$ on the \textsc{diff-Epitome} score. Trained with: \texttt{lr=1-e5, batch\_size=64, rank=1024}.}
    \label{fig:alphavsdiff}
\end{figure}

\paragraph{SFT} We observed a specialization effect provided by the SFT step, which improved two dimensions of \textsc{diff-Epitome} and BERTscore similarity. However, the more intensely the model is fine-tuned, the more it over-fits on the ED dataset, and the language understanding performance drops, as shown in Figure~\ref{fig:mmluvsalpha}. It is possible to mitigate this but not avoid it through careful hyperparameter optimization. First, as~\citealp{tunstall2023zephyr} suggests, we limited the training to one epoch. The key to reducing over-fitting was using a large \texttt{lora\_rank} ($r$) and low \texttt{lora\_alpha} ($\alpha$) relative to the rank. Rank controls the size of adapter matrices, and $\alpha$ scales the impact of the adapter weights on the base model. Large $\alpha$ supports retention of the original model's abilities, and rank $r$ controls the impact size of the adapter's fine-tuned weights. Figure~\ref{fig:alphavsdiff} shows the empathy metrics saturating with $\alpha\approx\frac{r}{4}$. Furthermore, we observed that increasing the rank above 1024 brings diminishing returns: the computational efficiency of LoRA vanishes while the over-fitting problem remains. We use the best configuration jointly for empathy and generalization ($r\!=\!1024, \alpha\!=\!64, lr\!=\!1e\!-\!5$) to report SFT results and subsequently to perform DPO training.

\begin{figure}[]
    \centering
    \includegraphics[width=\linewidth]{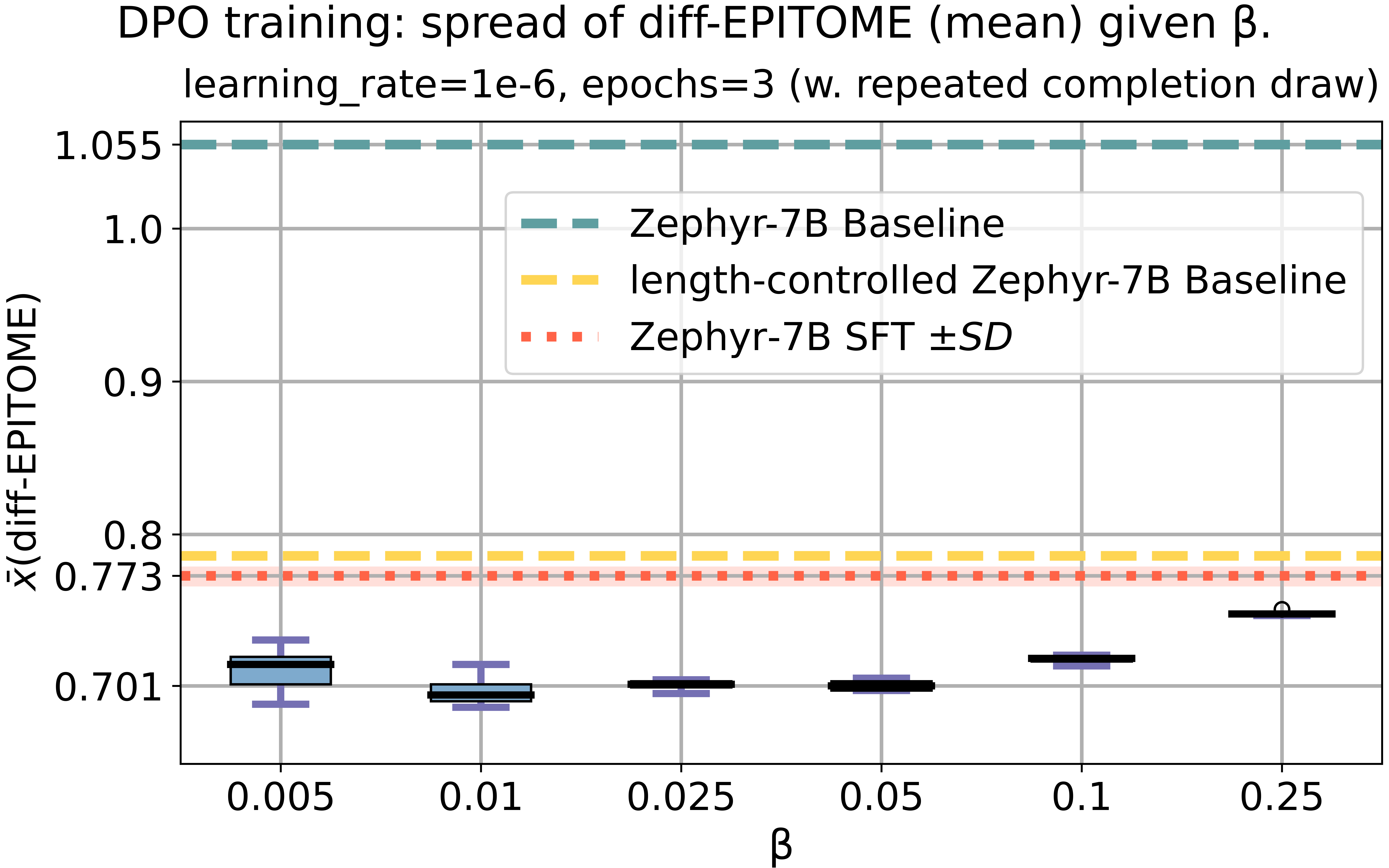}
    \caption{Stability of the DPO training measured by the spread of scores between multiple runs with the same hyperparameters, different seed and new completion draw; smaller spread signifying more stability. We measure the mean and spread of \textsc{diff-Epitome}, averaged across dimensions (ER, EX, IP): lower is better.}
    \label{fig:dpommludiff}
\end{figure}

\paragraph{DPO} The preference alignment with DPO improved the empathy metrics over the SFT while retaining the model's general performance. Nevertheless, its hyperparameters need to be set suitably to achieve this. Unsuitable hyperparameter configuration leads to training instability in addition to over-fitting (see Figure~\ref{fig:dpommludiff}). Notably, we faced problems with the stability of the training introduced by the dataset construction process: random selection of rejected completions from the group of dialogues labeled with polar opposite emotions. We solved it by lowering $\beta$ and learning rate and training for more epochs: re-drawing new randomly selected rejected completions for the same preferred ones for each epoch. The optimum configuration for high empathy scores while controlling over-fitting was training for three epochs, same as in~\citealp{tunstall2023zephyr}, with low $\beta$ ($0.05$) and moderate learning rate ($1e\!-\!6$).

\begin{figure}
    \centering
    \includegraphics[width=\linewidth]{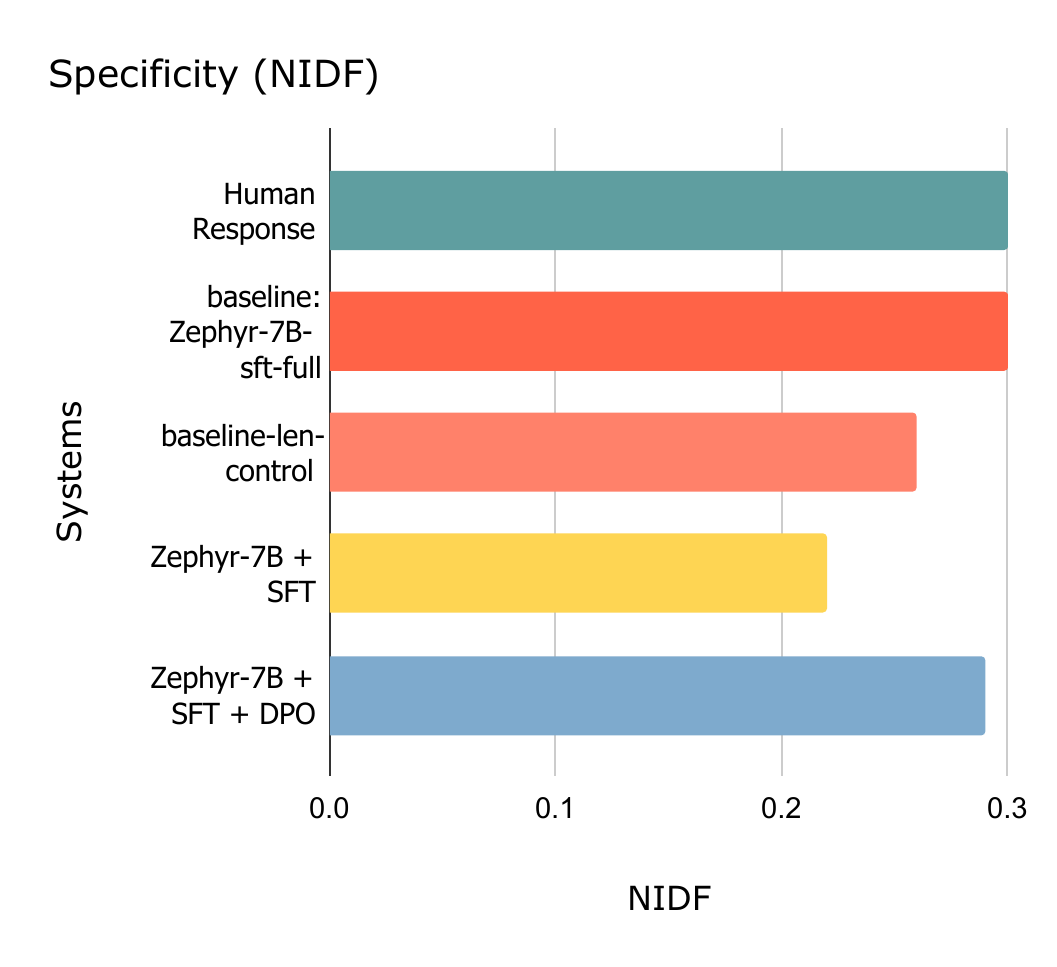}
    \caption{Evaluations based on \citet{lee-etal-2024-comparative}'s work: specificity with mean NIDF~\cite{see-etal-2019-makes}.}
    \label{fig:specificity-nidf}
\end{figure}

\begin{figure*}
    \centering
    \begin{tabular}{ccc}
      \includegraphics[width=.3\linewidth]{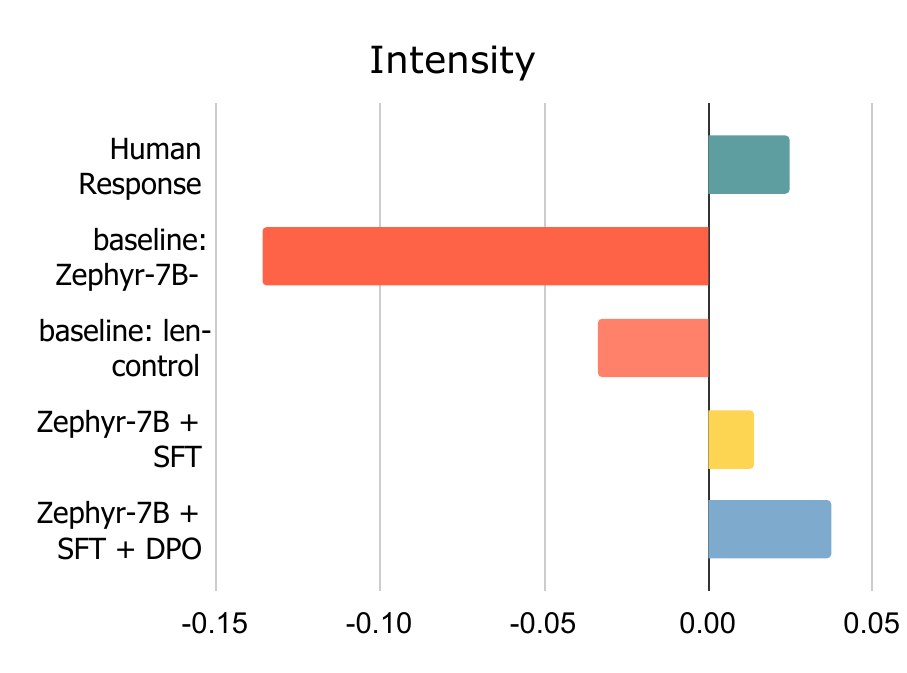}   &  \includegraphics[width=.3\linewidth]{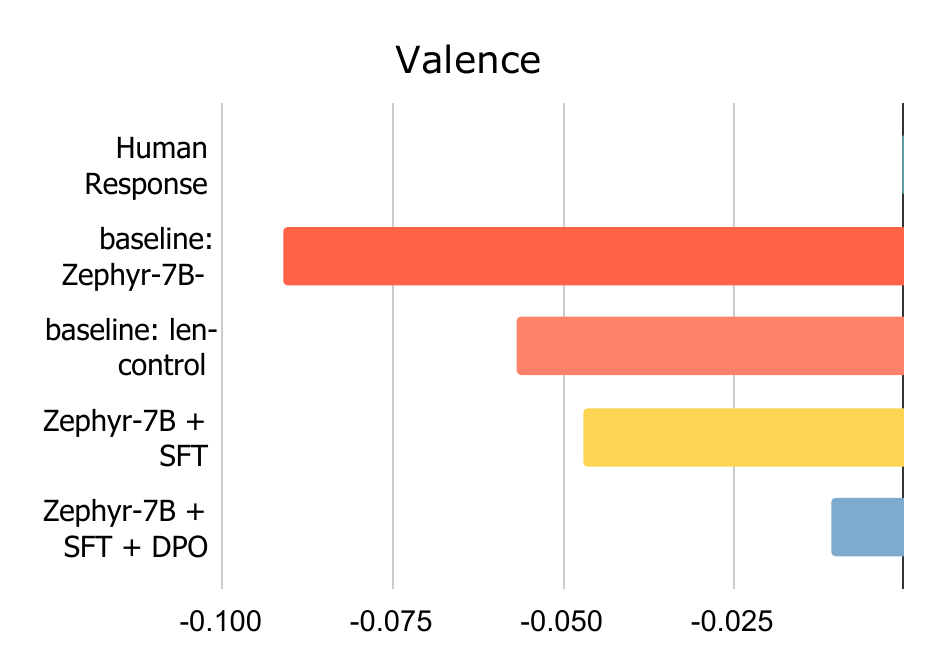} & \includegraphics[width=.3\linewidth]{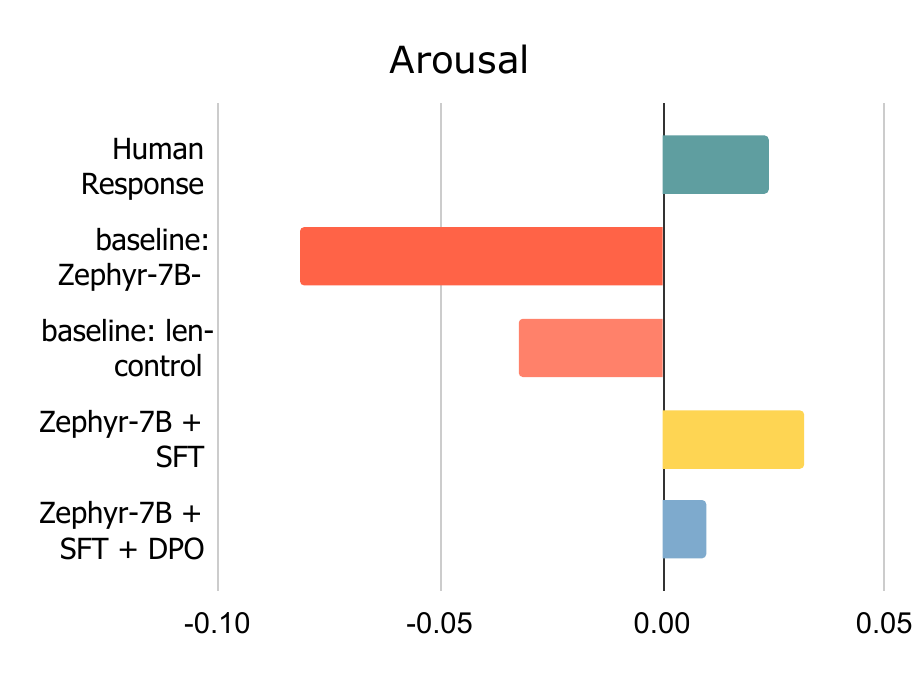} \\
    \end{tabular}
    \caption{Results of word choice metrics: intensity, valence, and arousal (IVA). Difference between the prompt and generated response score i.e., $\text{score}(\text{prompt}) - \text{score}(\text{utterance})$. The metric is further detailed in Appendix \ref{appx:vad_dists}.}
    \label{fig:vad_raw}
\end{figure*}

\begin{table*}[ht!]
  \centering
  \small
  \begin{tabular}{lccccc}
    \hline
    \textbf{Model} & \textbf{\# Templates$\uparrow$} & \makecell{\textbf{\# Span Nodes / } \\ \textbf{Total \# Nodes$\downarrow$}} & \makecell{\textbf{\# Children} \\ \textbf{From Root$\uparrow$}} & \makecell{\textbf{Compression} \\ \textbf{Ratio$\uparrow$}} & \makecell{\textbf{\# Unique} \\ \textbf{Start Words$\uparrow$}} \\
    \hline
    Human Response & 2526 (99.4\%) & \textbf{2746 / 22279 (12.3\%)} & \textbf{881 (34.7\%)} & \textbf{65.64\% }& \textbf{370} \\
    \hline
    \textit{baseline:} Zephyr-7B-sft-full & 2529 (99.6\%) &  5339 / 29080 (18.4\%) & 788 (31.0\%) & 44.53\% & 31 \\
    \textit{length-controlled baseline} &  2326 (91.6\%) &  2625/13425  (19.6\%) &  787 (31.0\%) & 53.33\% & 35 \\
    Zephyr-7B + SFT & 2390 (94.1\%) & 2085 / 9437 (22.1\%) & 741 (29.2\%) & 47.90\% & 48 \\
    Zephyr-7B + SFT + DPO & \textbf{2538 (99.9\%)} & \textbf{2927 / 20074  (14.6\%)} & \textbf{806 (31.7\%)} & \textbf{60.36\%} & \textbf{256}  \\

    \hline
  \end{tabular}
  \caption{Diversity metrics based on response-tries from \citet{lee-etal-2024-comparative}. \# Templates refers to the number of unique responses generated by each system. Compression ratio refers to the ratio between the size of tries made before and after the folding operations. A higher compression ration implies lower repetition. The metric is detailed in Appendix~\ref{appx:vad_dists}.}
  \label{tab:diversity}
\end{table*}

\paragraph{Diversity, Sensitivity, and Emotional Valence}
Across the diversity metrics shown in Table \ref{tab:diversity}, the human scores outperform all models, except in the case of \# Templates. DPO responses show the best results among the models for all other metrics. Generating more diverse responses is desirable because models tend to resort to generating generic responses like ``I am so sorry to hear that,'' which may bolster their performance on metrics that evaluate individual examples. \citet{lee-etal-2023-empathy} shows that \textsc{Epitome} and other representation-based models can take these shortcuts, which goes against the context-dependent nature of empathy.

Table \ref{tab:spec_vac} shows specificity, intensity, valence, and arousal. We performed Welch's t-tests to determine the significance of differences between each model and human values. For specificity, each model differs significantly from human responses, as shown in Figure~\ref{fig:specificity-nidf}. The Zephyr models have lower specificity ($t=-50.8$, $p<0.001$; $t=-2.1$, $p=0.039$) and baseline with higher ($t=2.6$, $p=0.012$). The models' specificities differ significantly from each other ($p<0.001$); SFT has the lowest and the baseline the highest, though there is an ideal point beyond which specificity does not improve model performance~\cite{Coll2017-lv}.
\begin{table}[h]
  \centering
  \small
  \addtolength{\tabcolsep}{-0.3em}
  \begin{tabular}{lc|lll}
    \hline
     & \textbf{Specificity} & \multicolumn{3}{c}{\textbf{Word choice}}  \\
    \textbf{Model}  & \textbf{NIDF$\uparrow$} & \textbf{I$\downarrow$} & \textbf{V$\downarrow$} & \textbf{A$\downarrow$}  \\
    \hline
    Human Response & $\mathbf{0.30}\pm0.07$ & $0.28$  & $0.14$ & $0.18$ \\
    \hline
        \textit{baseline:} Zephyr-7B& $\mathbf{0.30}\pm0.05$ & $\mathbf{0.24}$ * & $\mathbf{0.11}$ * & $\mathbf{0.15}$ * \\
    \textit{len-control baseline} & $0.26\pm0.05$ & $0.26$ * & $0.12$ * & $\mathbf{0.15}$ * \\
    SFT & $0.22\pm0.04$ & $0.29$ & $0.13$ * & $0.18$  \\
    SFT + DPO & $0.29\pm0.06$ & $0.27$ & $0.13 \ddagger$ & $0.16 \dagger$  \\

    \hline

    \hline
  \end{tabular}
  \caption{Evaluations based on \citet{lee-etal-2024-comparative}: specificity with mean NIDF and word choice (IVA). Values closer to 0 indicate the response had a closer IVA score to the prompt. * Indicates significant improvement over Human Response with $p<0.00001$, $\dagger$ with $p<0.001$, and $\ddagger$ with $p<0.01$ using permutation tests. The metric is detailed in Appendix~\ref{appx:vad_dists}.}
  \label{tab:spec_vac}
\end{table}

Figure~\ref{fig:vad_raw} shows the intensity-valance-arousal (IVA) metrics results. Desirable IVA metrics are closer to zero, signifying that IVA levels of responses are more similar to the prompts. The DPO model and baseline have significantly lower valence and arousal than humans, and the baseline has significantly lower intensity ($p < 0.05$). Between models, the baseline is significantly lower than the SFT and DPO models for all IVA metrics, and DPO is lower than SFT for arousal and intensity. Ultimately, the DPO model shows the most similar performance to human responses overall. For the absolute values of IVA metrics (as in \citet{lee-etal-2024-comparative}) and complete distributions of the metrics, see Appendix~\ref{appx:vad_dists}.

\subsection{Human Evaluation}
\begin{figure}[]
    \centering
    \includegraphics[width=\linewidth]{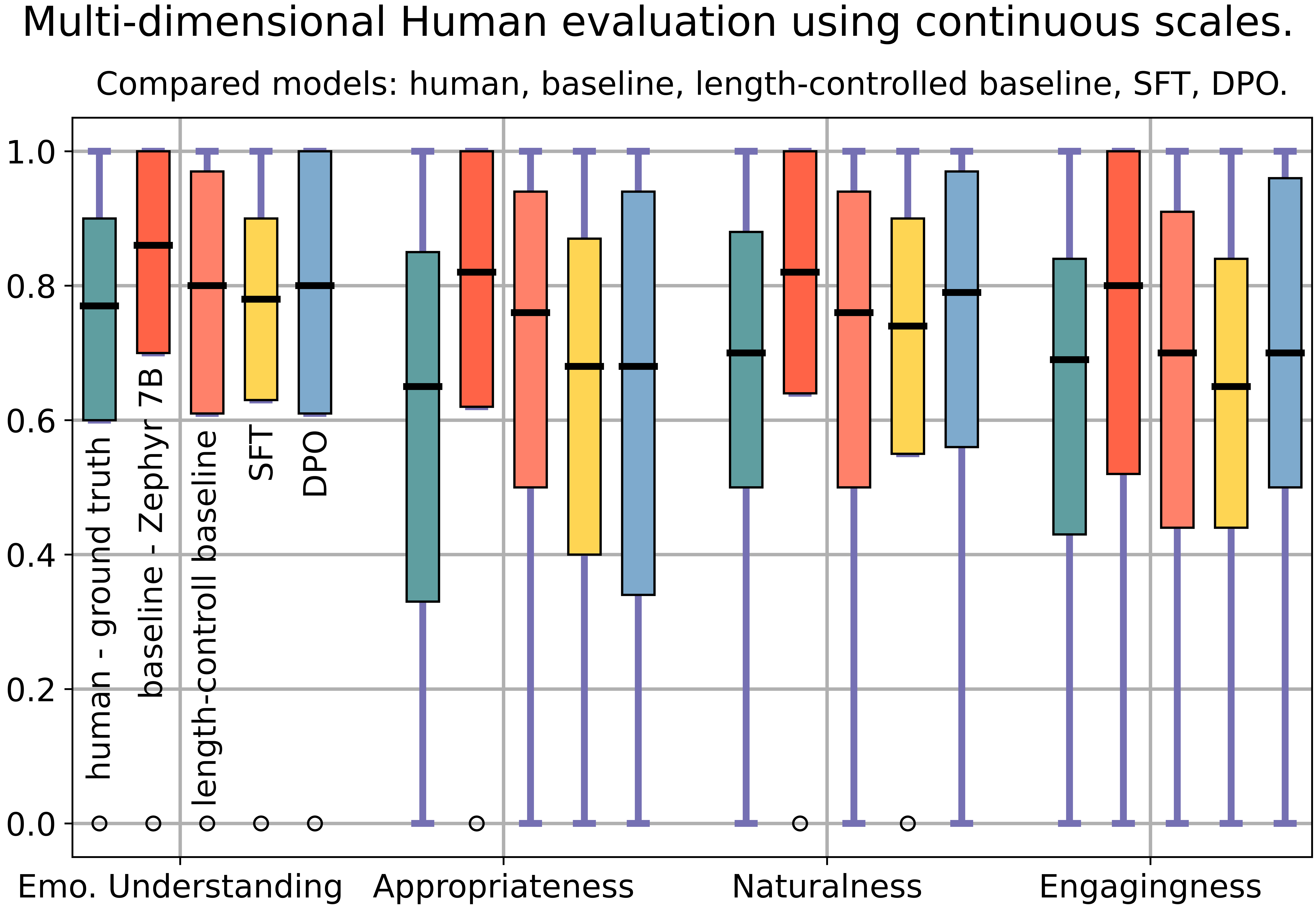}
    \caption{Distribution of human ratings along four dimensions of empathetic response using continuous scales (0-100). See Appendix~\ref{appx:HE} for the dimension descriptions.}
    \label{fig:humevalplot}
\end{figure}

Figure~\ref{fig:humevalplot} summarizes the comparative human evaluation. We evaluate samples from the \textsc{EmpatheticDialogues} corpus, and ask raters to evaluate empathetic engagement in four dimensions. Interestingly, samples directly taken from the corpus (\textit{human} in Fig.~\ref{fig:humevalplot}) were rated lower than their generated counterparts, while the Zephyr \textit{baseline} model was considered most empathetic. 

All of the models and the corpus sample share similar average scores in Consistency and Fluency ($\approx0.92$).


The DPO model was rated as equally or more empathetic than the length-controlled baseline and SFT models in most dimensions and consistently scored higher in \textit{response naturalness}. However, the baseline outperformed the DPO and SFT models in \textit{situational appropriateness}, likely due to a learned bias from fine-tuning by focusing on 'situational empathy' in the ED corpus. However, the difference was small but statistically significant.

The ratings are relatively independent of the emotion label that they were sampled from, and dimension ratings are not correlated. In summary, across independent dimensions of empathy, human raters showed a mixed preference leaning toward the DPO model's responses (excluding the Zephyr baseline, which was overwhelmingly preferred). Consistently low ratings across empathy dimensions for samples from the original corpus also echo concerns about corpus quality from~\cite {debnath-2023-critical}.

\subsection{Response Length: Observations}
The mean length of ground truth and the DPO responses is 65 and 63 characters respectively, 55 for SFT, and 152 for the Zephyr-7B baseline. The long-response bias affects each of the automated metrics and also affects the human evaluation, which we were unable to fully mitigate by the survey design. This is not surprising, as there is evidence of cognitive bias in raters' assessments of qualitative statements about language based on the length of the responses~\citep{santhanam2020studying,park2024disentangling}. Therefore, we introduced a control by suggesting shorter messages in the system prompt and the \texttt{max\_tokens} parameter of the generation pipeline, resulting in the \textit{length-controlled baseline} with mean length of 75 characters.  Interestingly, only some of the emotion scores in the human evaluation dimensions correlate with the generated outputs' length. See Table~\ref{tab:eff_sizes} for detailed analysis of effect sizes.



\section{Conclusion}
\label{sec:conclusion}
We have successfully aligned large language models (LLMs) for empathetic response generation using preference datasets and Direct Preference Optimization (DPO). Our approach demonstrated that preference data can be effectively constructed using Plutchik’s polar emotion model. Through extensive experimentation, we highlighted how varying hyperparameters can mitigate the specialization effect of training and influence empathy metrics. Our findings, validated through the \textsc{EmpatheticDialogues} dataset, \textsc{diff-Epitome}, BERTscore, and diversity and emotion valence automated metrics, as well as human evaluations, provide a solid foundation for further advancements in the preference-based alignment of LLMs for empathetic response generation.

\section*{Limitations}
\label{sec:limitations}
While our method certainly improves the congruence of responses with the ground truth, we believe that training solely on the \textsc{EmpatheticDialogues} dataset alone is not going to fulfill the ultimate goal of aligning LLMs for ERG: generating more empathetic responses in all domains. However, we believe that we have shown the potential of polar emotion models to guide preference dataset generation, on which future work can build.

In human evaluation, we have encountered the well-documented human preference for longer responses~\cite {santhanam2020studying,park2024disentangling}, which should not be disregarded. Future work should, therefore, focus on combining both insights from our work: collecting or generating long responses while guiding the preference dataset construction using the polar emotion models.



\bibliography{custom}
\bibliographystyle{acl_natbib}

\appendix

\section{Training Details}
\label{appx:trainingsft}\label{appx:trainingdpo}
\paragraph{Code} {\small \url{github.com/justtherightsize/empo}}

\paragraph{Models}
\begin{itemize}[leftmargin=0pt,parsep=0pt]
    \item[] \textit{baseline}: 
    \begin{itemize}[leftmargin=*,parsep=0pt]
        \small
        \item[] \url{hf.co/alignment-handbook/zephyr-7b-sft-full}
    \end{itemize}
    \item[] SFT: 
    \begin{itemize}[leftmargin=*,parsep=0pt]
        \small
        \item[]\url{hf.co/justtherightsize/zephyr-7b-sft-full124}
    \end{itemize}
    \item[] DPO: 
    \begin{itemize}[leftmargin=*,parsep=0pt]
        \small
        \item[]\url{hf.co/justtherightsize/zephyr-7b-sft-full124_d270}
    \end{itemize}
\end{itemize}
See the corresponding model cards for usage.

\paragraph{Training Settings}
We trained the models using the HuggingFace Transformers, TRL, and PEFT libraries. In the SFT step, we applied standard sequence-to-sequence training with \textbf{cross-entropy} loss on tokens from the user prompts (not on the tokens from "replies"). In both the SFT and DPO training steps, we optimized the model with \textbf{AdamW} optimizer and \textbf{batch size} of 64. We used \textbf{learning rate} of 1e-5 for the SFT and 1e-6 for DPO without \textbf{warmup} and with a cosine \textbf{lr decay}. The models were trained in bf16 \textbf{precision}. 

\paragraph{Hardware}
To train our models, we used three NVIDIA A100 80GB GPUs and four A100 40GB GPUs. We trained LoRA adapters with ranks ranging from 16 to 2048 for models with 7B parameters. The total training wall time, including preliminary experiments, was 33 days.

\paragraph{\textsc{EmpatheticDialogues}}
We use the version from \url{hf.co/datasets/empathetic_dialogues} filtering out $<1\%$ malformed dialogues by keywords. For details we refer the reader to the code repository functions \texttt{ed\_load.\string{load\_preprocess\_ed, get\_ed\_for\_generation, get\_ed\_for\_dpo\string}}.

\paragraph{DPO Preference Dataset\label{appx:plutchik}}
For training with DPO, we introduced a method to construct a preference dataset from the \textsc{EmpatheticDialogues} dataset. With each training run or a single epoch within a multi-epoch run, the dataset is created anew as described in Section~\ref{sec:experiments}. This section contains auxiliary material to support the description of the preference dataset creation. 

\begin{figure}[h]
    \centering
    \includegraphics[width=.85\linewidth]{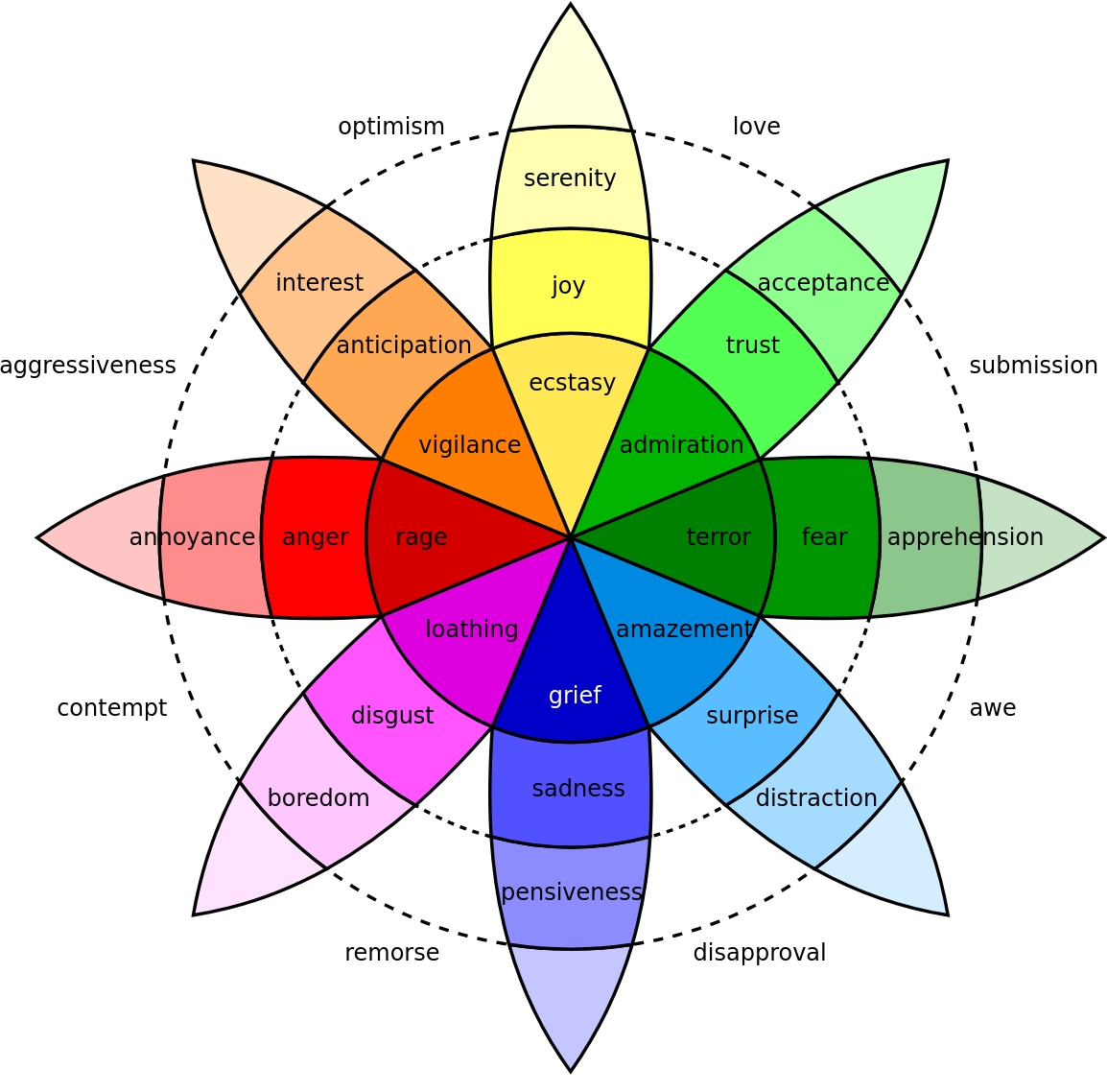}
    \caption{Plutchik's wheel of emotions (Creative Commons). It shows eight basic emotions: joy, trust, fear, surprise, sadness, anticipation, anger, and disgust. The wheel of emotions groups these eight basic emotions based on the physiological purpose of each into polar coordinates reflecting their similarity and intensity.}
    \label{fig:plutchik}
\end{figure}

\begin{figure}[h]
    \centering
    \includegraphics[width=.8\linewidth]{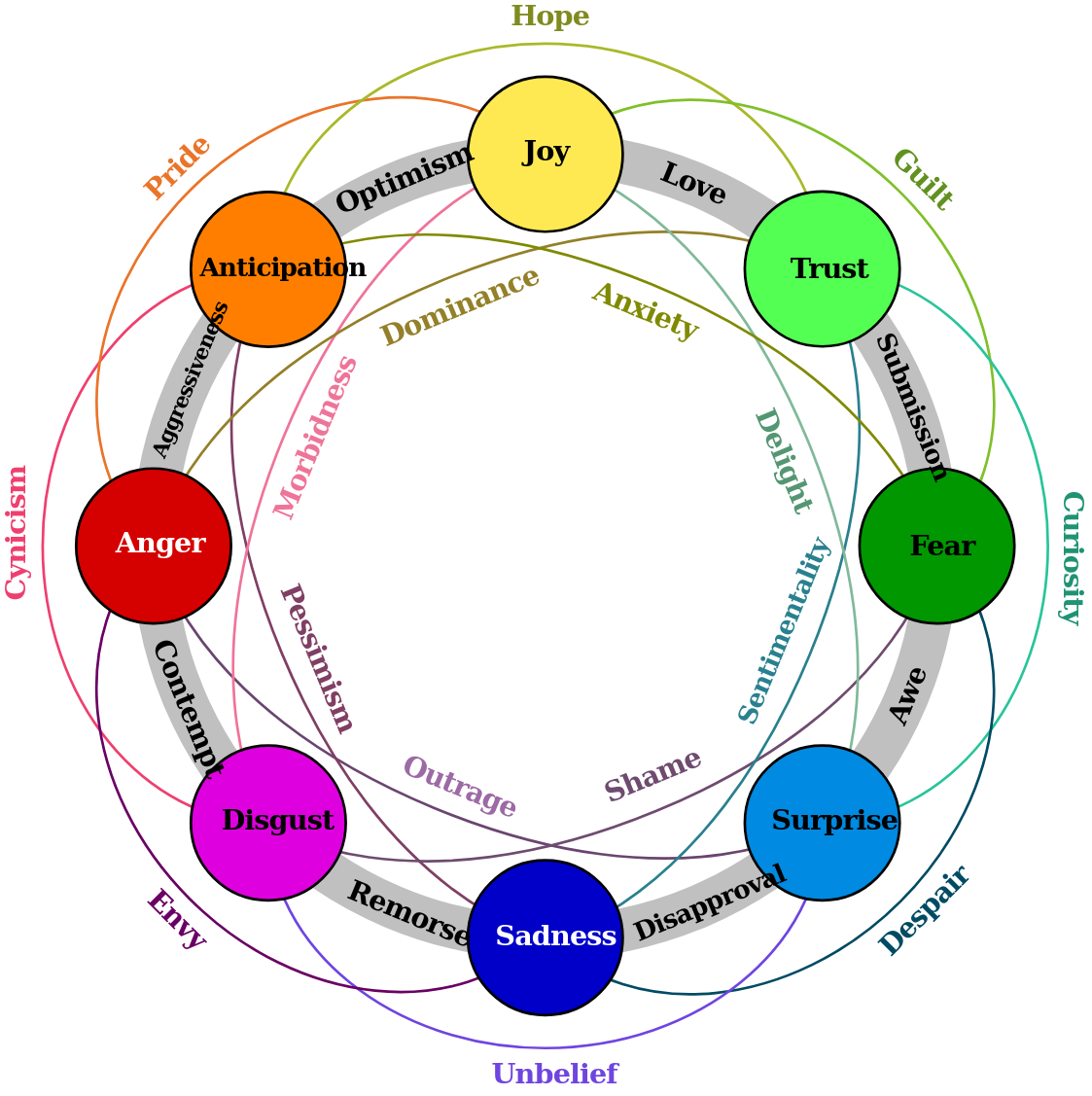}
    \caption{Plutchik's emotion's dyads (Creative Commons). When two emotions are elicited together, they form the primary dyad. If they are one petal apart, it is a secondary dyad; if they are two petals distant from each other, it is a tertiary dyad. Opposite dyads are on the opposite side.}
    \label{fig:dyads}
\end{figure}

Each dialog in the \textsc{EmpatheticDialogues} dataset is associated with an emotion label. To construct the training example for DPO, we paired the preferred dialog completion (in our case, using the ground truth) with one rejected competition. To find the rejected completion, we proposed using opposite emotion labels. The opposites are based on two theory-based sources: Plutchik's wheel (Figure~\ref{fig:plutchik}), emotional dyads (Figure~\ref{fig:dyads}), and we proposed the rest ourselves. The opposites form the lookup Table~\ref{tab:opposites}, which is queried each time a preferred/rejected pair is constructed.

\begin{table}[ht!]
  \centering
  \begin{tabular}{lll}
  \hline
    \textbf{Emotion label}      &\textbf{Opposite label}    &\textbf{Source}\\ \hline
    afraid                      &angry&wheel\\
    angry                       &afraid&wheel\\
    sad                         &joyful&wheel\\
    grateful                    &disgusted&wheel\\
    surprised                   &anticipating&wheel\\
    trusting                    &disgusted&wheel\\
    disgusted                   &trusting&wheel\\
    anticipating                &surprised&wheel\\
    content                     &anxious&wheel\\
    apprehensive                &annoyed&wheel\\
    joyful                      &sad&wheel\\\hline
    proud                       &ashamed&dyads\\
    prepared                    &anxious&dyads\\
    ashamed                     &proud&dyads\\
    guilty                      &proud&dyads\\
    nostalgic                   &hopeful&dyads\\
    anxious                     &content&dyads\\
    hopeful                     &nostalgic&dyads\\\hline
    sentimental                 &apprehensive&\\
    jealous                     &faithful&\\
    embarrassed                 &confident&\\
    excited                     &devastated&\\
    annoyed                     &apprehensive&\\
    lonely                      &caring&\\
    faithful                    &jealous&\\
    terrified                   &furious&\\
    confident                   &embarrassed&\\
    furious                     &terrified&\\
    disappointed                &impressed&\\
    caring                      &lonely&\\
    impressed                   &disappointed&\\
    devastated                  &excited&\\\hline
  \end{tabular}
  \caption{The opposite emotion labels lookup table. The emotion labels are sourced from the \textsc{EmpatheticDialogues} dataset. Each pair of opposites is associated with how the pair was determined: using Plutchiks's wheel, emotion dyads, or our proposal.}
  \label{tab:opposites}
\end{table}

\section{Open LLM Leaderboard Tasks\label{appx:openllm}}
Here we describe the tasks what make up the Open LLM leaderboard based on the  documentation~\cite{open-llm-leaderboard-tasks}.

\paragraph{Big Bench Hard (BBH)}~\cite{suzgun-etal-2023-challenging} is a subset of 23 challenging tasks from the BigBench dataset to evaluate language models. The tasks use objective metrics, are highly difficult, and have sufficient sample sizes for statistical significance. They include multistep arithmetic, algorithmic reasoning (e.g., boolean expressions, SVG shapes), language understanding (e.g., sarcasm detection, name disambiguation), and world knowledge. BBH performance correlates well with human preferences, providing valuable insights into model capabilities.
\paragraph{MATH}~\cite{hendrycksmath2021} is a compilation of high-school level competition problems gathered from several sources, formatted consistently. Generations must fit a very specific output format. A subset of level 5 MATH questions only is included in the benchmark.
\paragraph{Graduate-Level Google-Proof Q\&A Benchmark (GPQA)}~\cite{rein2023gpqagraduatelevelgoogleproofqa} is a highly challenging knowledge dataset with questions crafted by PhD-level domain experts in fields like biology, physics, and chemistry. These questions are designed to be difficult for laypersons but relatively easy for experts. The dataset has undergone multiple rounds of validation to ensure both difficulty and factual accuracy. Access to GPQA is restricted through gating mechanisms to minimize the risk of data contamination.
\paragraph{Multistep Soft Reasoning (MuSR)}~\cite{osti_10503204} is a dataset consisting of algorithmically generated complex problems, each around 1,000 words in length. The problems include murder mysteries, object placement questions, and team allocation optimizations. Solving these problems requires models to integrate reasoning with long-range context parsing. Few models achieve better than random performance on this dataset.
\paragraph{Massive Multitask Language Understanding - Professional (MMLU-Pro)}~\cite{wang2024mmluprorobustchallengingmultitask} is a refined version of the MMLU~\cite{hendrycks2020measuring} dataset, which has been a standard for multiple-choice knowledge assessment. Recent research identified issues with the original MMLU, such as noisy data (some unanswerable questions) and decreasing difficulty due to advances in model capabilities and increased data contamination. MMLU-Pro addresses these issues by presenting models with 10 choices instead of 4, requiring reasoning on more questions, and undergoing expert review to reduce noise. As a result, MMLU-Pro is of higher quality and currently more challenging than the original.
\paragraph{IFEval}~\cite{zhou2023instruction} is a dataset designed to test a model’s ability to follow explicit instructions, such as "include keyword x" or "use format y". The focus is on the model’s adherence to formatting instructions rather than the content generated, allowing for the use of strict and rigorous metrics.

\section{Automatic Empathy Evaluation: Definitions and Implementation}
\label{appx:vad_dists}

In this section, we detail the automatic empathy evaluation methods, including a brief definition, method of evaluation, and some discussion on the impact of the metric on the intepretation of the results. We used automatic empathy evaluation methods detailed in \citet{lee-etal-2024-comparative}, with the exception of the PAIR metric \citep{min-etal-2022-pair}, which we were not able to reliably reproduce.

\paragraph{Specificity} In the patient-practitioner context, specificity is defined as the degree to which the practitioner comments on the generalities versus specific emotions. \citet{lee-etal-2024-comparative}'s approach uses normalized inverse document frequency (NIDF) to determine the specificity of a dialogue, as it is a normalized ratio of how rare a word is in a given text.

\paragraph{Word Choice} When evaluating empathy, language which is both rich and consistent with the conversant's discourse is perceived to be more empathetic. The deviation within the dimensional aspects of emotion (intensity, valence, and arousal) between interactions is used as a metric to gauge the emotional consistentcy in word choice. The NRC Emotion Intensity and NRC-VAD Lexicons were used as the reference.

\begin{figure*}
    \centering
    \begin{tabular}{ccc}
      \includegraphics[width=.3\linewidth]{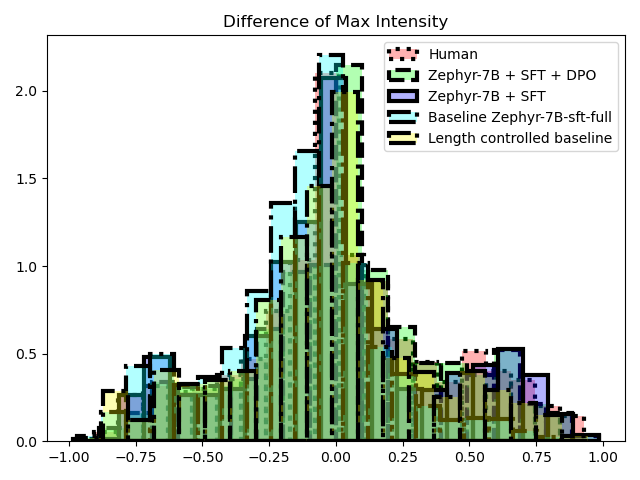}   &  \includegraphics[width=.3\linewidth]{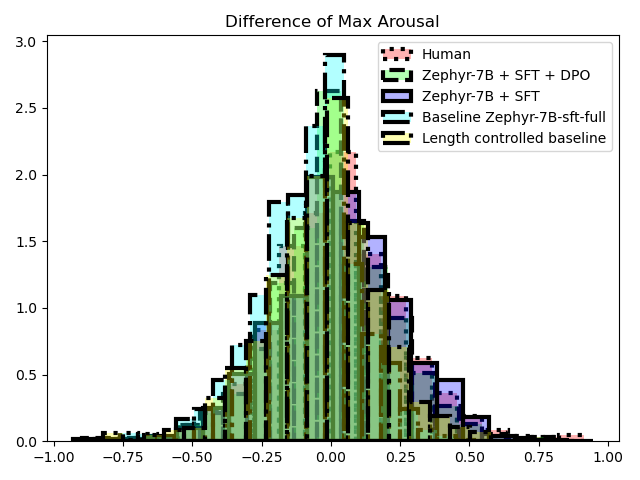} & \includegraphics[width=.3\linewidth]{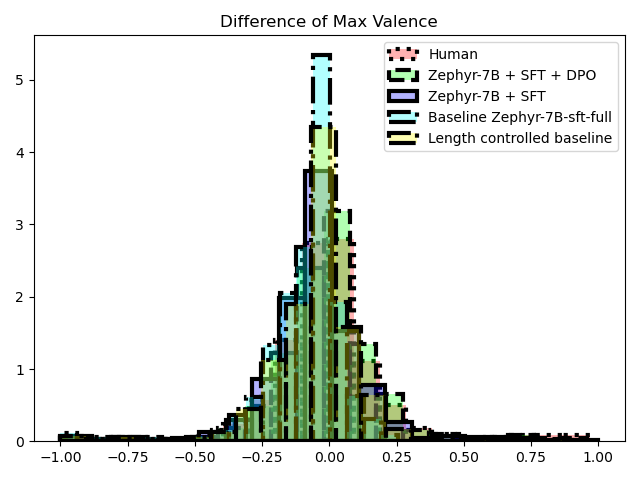} \\
    \end{tabular}
    \caption{Valence, intensity, and arousal distributions for the two models, two baselines, and human responses.}
    \label{fig:vad_diffs}
\end{figure*}

Here we present the distributions of valence, arousal, and intensity across our models from \S\ref{sec:experiments}. A distribution for each metric is shown in Figure~\ref{fig:vad_diffs}. We provide this distribution to highlight that there are two possible intepretations of the IVA deviation metric: the absolute distance or the signed difference. \citet{lee-etal-2024-comparative} adopts the former approach.

\begin{figure*}
    \centering
    \begin{tabular}{ccc}
      \includegraphics[width=.3\linewidth]{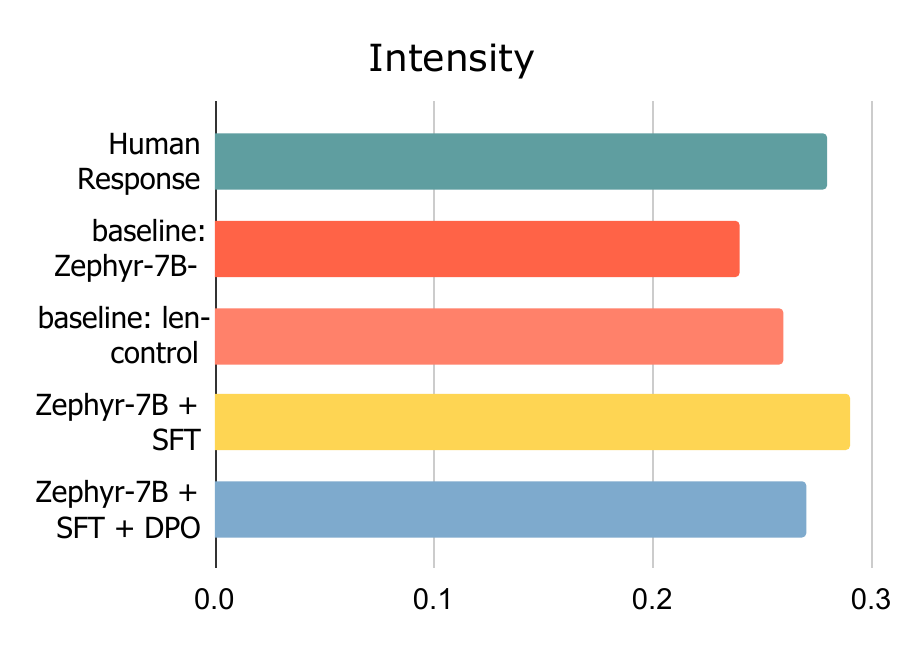}   &  \includegraphics[width=.3\linewidth]{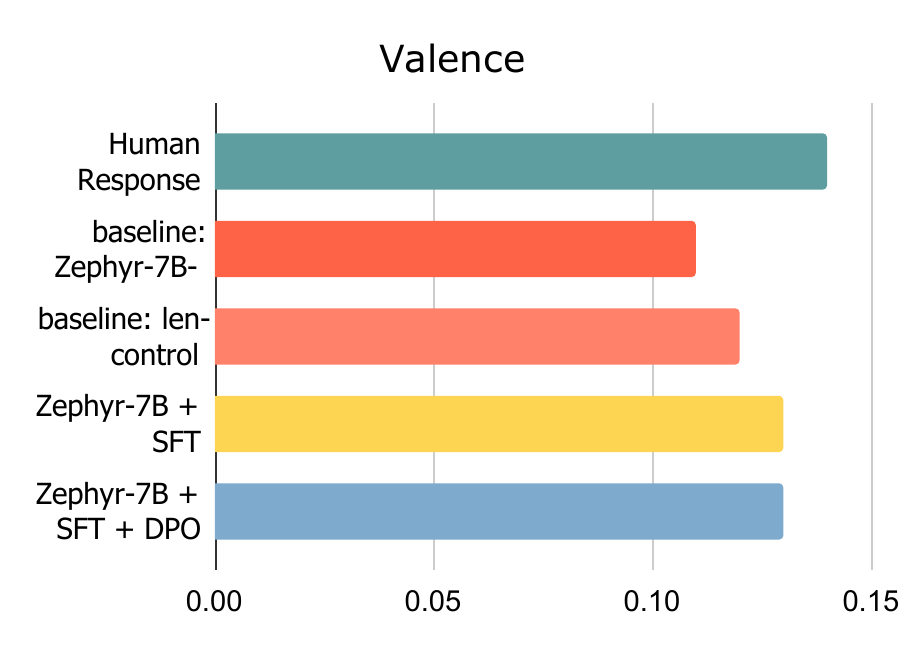} & \includegraphics[width=.3\linewidth]{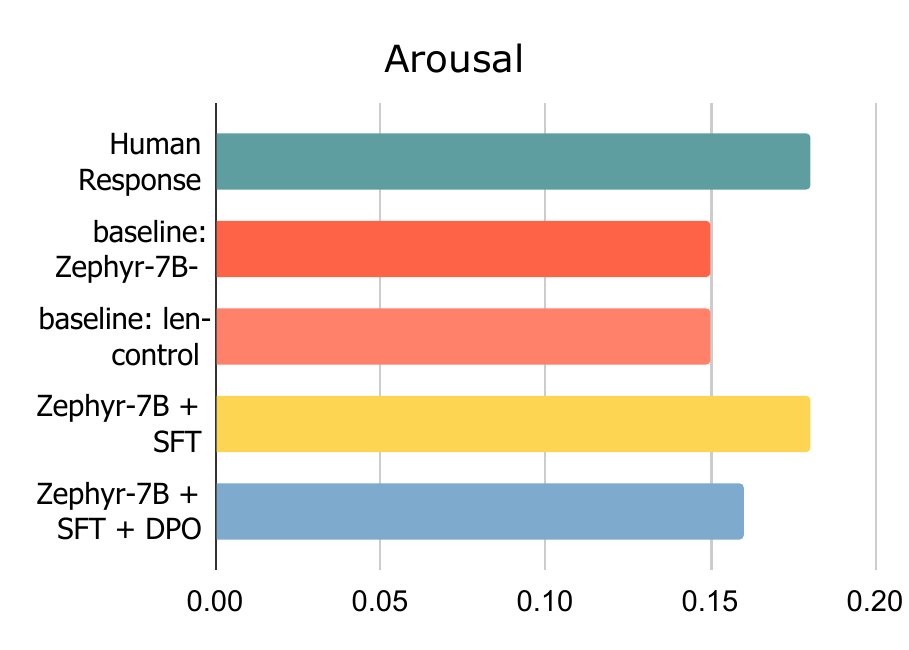} \\
    \end{tabular}
    \caption{Results of word choice metrics: intensity, valence, and arousal. Distances between the prompt and generated response utterance scores for emotional intensity, valence, and arousal, i.e., $|\text{score}(\text{prompt}) - \text{score}(\text{response})|$.}
    \label{fig:vad_abs}
\end{figure*}

Furthemore, we present the absolute values of the IVA metrics from \S\ref{sec: results} in Figure \ref{fig:vad_abs}. The signed difference approach provides more detailed insight into emotion patterns in word choice, in addition to those presented in the main body of the paper.

\paragraph{Diversity} Instrumental to the generalizability of empathy evaluation, diversity aims to examine a mode's capability to generate new responses based on a single utterance, so as to ensure that the model is generating distinct responses rather than a set of templates. \citet{lee-etal-2024-comparative} develops a method to examine diversity based on response-tries, i.e. a folded token-bu-token trie representation of the responses provided by the model on repeated prompts. These response tries are constructed by ``folding'' over a response whenever a unique token is observed. Once prompted for a test set, the properties of the constructed tries, such as the number of templates (i.e. the number of distinct response structures) and compression ratio (the ratio of the size of the response tries before and after folding).

\section{Design of Human Evaluation Survey}\label{appx:HE}

Human evaluation of EmPO was done on the Prolific platform~\citep{prolific}, by annotators screened for their education level and experience in Psychology and Social Work.

\begin{table*}[h]
  \centering
  \small
  \begin{tabular}{l|l|llll}
    \hline
	 & & \multicolumn{4}{c}{\textbf{Cohen's d}} \\
      & & \textbf{Emotion} & \textbf{Situational} & \textbf{Contextual} &\textbf{Conversational}\\
     \textbf{Output source}  & \textbf{Greater score than} & \textbf{Understanding} & \textbf{Appropriateness} & \textbf{Naturalness} &\textbf{Engagingness}\\
     \hline
    \textit{baseline:}& Human targets & $ 0.58$* & $ 0.88$* & $ 0.48$* & $ 0.47$* \\
                      & baseline-len-control & $ 0.39$* & $ 0.42\dagger$ & $ 0.3\ddagger$ & $ 0.45$* \\
                      & SFT & $ 0.5$* & $ 0.72$* & $ 0.42\dagger$ & $ 0.53$* \\
                      & SFT + DPO & $ 0.38\dagger$ & $ 0.7$* & $ 0.17$ & $ 0.3\dagger$ \\
    \hline
    SFT + DPO & Human targets & $ 0.18$ & $ 0.14$ & $ 0.32\ddagger$ & $ 0.18\ddagger$ \\
              & baseline-len-control & $-0.04$ & $ 0.25$* & $-0.16$ & $-0.17$ \\
              & SFT & $ 0.2$ & $ 0$ & $ 0.25\ddagger$ & $ 0.22\ddagger$  \\
    \hline
    baseline-len-control & Human targets & $ 0.14$ & $ 0.41\dagger$ & $ 0.15$ & $0$ \\
                         & SFT & $ 0.06$ & $ 0.26\ddagger$ & $ 0.08$ & $ 0.04$\\
    \hline
    SFT & Human targets & $ 0.19$ & $ 0.01$ & $ 0.01$ & $-0.01$ \\
    \hline
    
  \end{tabular}
  \caption{We used a t-test to test a one-sided hypothesis that the annotators rate some model outputs (left) higher than others (right), * with $p<0.00001$, $\dagger$ with $p<0.001$, and $\ddagger$ with $p<0.05$. After the Bonferronni correction most of the results are not significant on $\alpha = 0.05$. But the difference between the corpus-given targets and the \textit{baseline} is suggesting a bias in the survey design, sub-optimal empathy decomposition coupled with non-ideal dimension definitions, or an issue with the \textsc{EmpatheticDialogues} dataset itself as suggested by~\citet{debnath-2023-critical}.}
  \label{tab:eff_sizes}
\end{table*} 

Existing human evaluations are often done on a discrete Likert scale, along with A/B testing of responses (see Table~\ref{tab:rw-eval}). Current benchmark evaluations use a three-part test of \textit{relevance}, \textit{fluency}, and \textit{empathy/sympathy} \citet{rashkin-etal-2019-towards}. As this neither expounds on the multifaceted nature of empathy, nor does it provide a tractable comparison for models, extant literature has adopted several other facets considered important in an empathetic conversational agent (i.e. informativeness, safety, etc.). For our work, we deemed A/B testing unsuitable, as the models generate semantically and situationally similar responses, inducing the cognitive bias towards preferring sentences of longer lengths as opposed to the more ``empathetic'' ones. We also preferred a continuous 0-100 rating scale in order to maintain tractability \cite{ji-etal-2022-achieving}.

We use a four-dimensional framework to evaluate empathy in the language, i.e. the design was chosen to best reflect the perceived empathy of the model based on a single conversation. In Table \ref{tab:he-parameters}, we detail the parameters, displayed statements, and an examination of its choice as a parameter.

As emotions, empathy, and conversation are dynamic, a static evaluation is the aggregate perception of the annotators' examination of a single instance of a conversation. Therefore, questions about the nature of the model were avoided in favour of response about the nature of the response, as their rating of empathetic measures have to be explicitly based on the language generated by the model and any cues therein. 

The human evaluation included two pilot studies. Both pilot studies included 90 participants, recruited using Prolific under the same constraints as the final human evaluations. Finally, the ground truth responses were evaluated as a control.

When extracting examples for the evaluation we first grouped them, then randomly sampled four from each. The conversations were shown to annotators in the same order, but we block-randomized the samples, so that some annotators evaluated outputs of more than one model, in batches of 15. For DPO, base models and human targets, the randomization scheme is as follows: block the outputs, so that 5 examples in a row are from the same model resulting in $3!=6$ orderings. Adding $3$ batches that evaluate samples of one model, making each sample annotated three times. For the length-controlled baseline, we skipped the block-randomization.

\begin{figure*}
    \centering
    \includegraphics[width=0.80\linewidth]{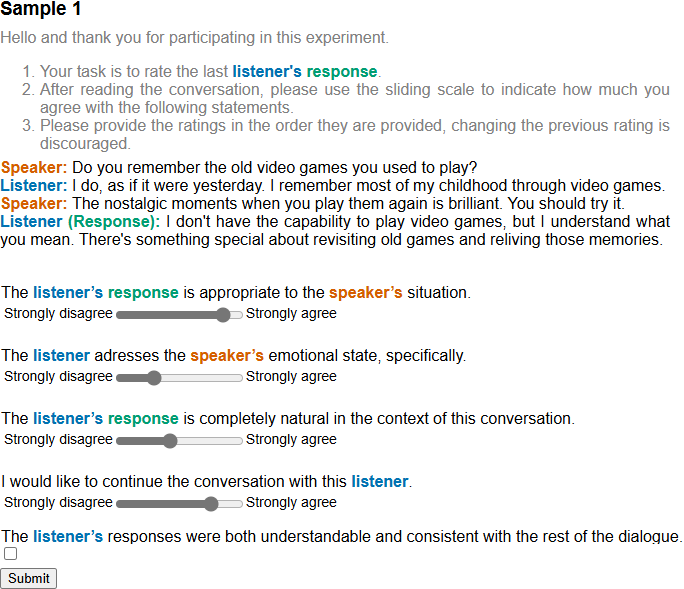}
    \caption{The evaluation page for annotation. Annotators were additionally shown an introductory page with in-depth instructions and a solved example with the same structure.}
    \label{fig:he_page_example}
\end{figure*}

\begin{table*}[t]
    \centering
    \begin{tabularx}{\textwidth}{ X|X|X }
       \bf Parameter & \bf Statement & \bf Explanation \\ \hline
       
        \textbf{Emotion Understanding} & The \textit{listener}'s response addresses the \textit{speaker}'s emotional state, specifically & Emotion understanding is essential for empathetic engagement, but the agent's comprehension of the user's emotions can only be assessed if the response explicitly addresses them.  \\\hline

        S\textbf{ituational Appropriateness} & The \textit{listener}’s response is appropriate to the \textit{speaker}’s situation. &We examine situational appropriateness because it allows for the consideration of context in empathetic responses, unlike a single prompt. Additionally, the ED corpus uses a situational model where conversations are linked to specific emotions and situations. Thus, we assess how well the annotator perceives the agent's understanding of the situation.\\\hline

        \textbf{Contextual Naturalness} & The \textit{listener}'s response is completely natural in the context of this conversation. & Human conversation reflects understanding of emotional and situational context through the nature and tone of responses. This question aims to assess how closely the agent's response mirrors a human response in a similar context.\\\hline

        \textbf{Conversational Engagingness} & I would like to continue the conversation with this listener. & This question focuses on the user's willingness to continue the conversation with the agent, rather than the conversation itself. Given empathy's role in fostering user-agent bonding, we assess whether annotators would like to continue. However, unwillingness may stem from factors beyond the agent's engagingness.\\\hline

        \textbf{Consistency \& Fluency} & The \textit{listener}’s responses were both understandable and consistent with the rest of the dialogue \textit{(yes/no)}. & Relevance and fluency are a staple part of dialogue evaluation. However, both with the advancement of LLMs in general, and the similar base model for our examination in particular, model responses were consistently found to be both consistent and fluent, through both pilots. \\\hline
    \end{tabularx}
    \caption{The parameters of human evaluation with their statement (as provided to the annotators when given instructions), and a brief explanation of why the parameter was selected and the rationale for framing the evaluation statements.}
    \label{tab:he-parameters}
\end{table*}

\end{document}